\documentclass[conference]{IEEEtran}
\IEEEoverridecommandlockouts
\usepackage{cite}
\usepackage{amsmath,amssymb,amsfonts}
\usepackage{algorithm}
\usepackage{graphicx}
\usepackage{textcomp}
\usepackage{xcolor}
\usepackage[noend]{algpseudocode}
\usepackage{subfig}
\usepackage{hhline}
\usepackage{multirow}
\usepackage{hyperref}

\def\BibTeX{{\rm B\kern-.05em{\sc i\kern-.025em b}\kern-.08em
    T\kern-.1667em\lower.7ex\hbox{E}\kern-.125emX}}
\begin{document}

\title{ACO based Adaptive RBFN Control for Robot Manipulators\\
}

\author{\IEEEauthorblockN{Sheheeda Manakkadu}
\IEEEauthorblockA{\textit{Computer and Information Science} \\
\textit{Gannon University}\\
Erie, PA \\
mariamma001@gannon.edu}
\and
\IEEEauthorblockN{Sourav Dutta}
\IEEEauthorblockA{\textit{Computer Science} \\
\textit{Ramapo College of New Jersey}\\
Mahwah, NJ \\
sdutta1@ramapo.edu}
%\and
%\IEEEauthorblockN{Mei-Huei Tang}
%\IEEEauthorblockA{\textit{Software Engineering} \\
%\textit{Gannon University}\\
%Erie, PA \\
%tang002@gannon.edu}
%\and
%\and
%\and
%\and
}

\maketitle

\begin{abstract}
This paper describes a new approach for approximating the inverse kinematics of a manipulator using an Ant Colony Optimization (ACO) based RBFN (Radial Basis Function Network). In this paper, a training solution using the ACO and the LMS (Least Mean Square) algorithm is presented in a two-phase training procedure. To settle the problem that the cluster results of k-mean clustering Radial Basis Function (RBF) are easy to be influenced by the selection of initial characters and converge to a local minimum, Ant Colony Optimization (ACO) for the RBF neural networks which will optimize the center of RBF neural networks and reduce the number of the hidden layer neurons nodes is presented. Compared with k-means clustering RBF Algorithm, the result demonstrates that the accuracy of Ant Colony Optimization for the Radial Basis Function (RBF) neural networks is higher, and the extent of fitting has been improved.     
\end{abstract}


\section{INTRODUCTION}

	In robot kinematics, there are two important problems, forward and inverse kinematics. Forward kinematics can be regarded as a one-to-one mapping from the joint variable space to the Cartesian coordinate space (world space). From a set of joint angles, forward kinematics determines the corresponding location (position and orientation) of the end-effecter. This problem can be easily solved by the $4 x 4$ homogenous transformation matrices using the Denavit \& Hartenbergh representation \cite{1}. Inverse kinematics is used to compute the corresponding joint angles from the location of the end-effecter in space. Obviously, inverse kinematics is a more difficult problem than forward kinematics because of its multi-mapping characteristic. There are many solutions to solve the inverse kinematics problem, such as the geometric, algebraic, and numerical iterative methods. In particular, some of the most popular methods are mainly based on the inversion of the mapping established between the joint space and the task space by the Jacobian matrix \cite{2}. This solution uses numerical iteration to invert the forward kinematics Jacobian matrix and does not always guarantee to produce all the possible inverse kinematics solutions and involves significant computation. In cases where the manipulator geometry cannot be exactly specified, the traditional methods become very difficult or impossible, for example, the robot-vision system. The artificial neural network, which has significant flexibility and learning ability, has been used in many robot control problems. In fact, for the inverse kinematics problem, several neural network architectures have been used, such as MLPN (Multi-Layer Perceptron Network), Kohonen self-organizing map, and RBFN. In \cite{3,4} Guez et al and Choi described solutions using the MLPN and backpropagation training algorithm. Additionally, Watanabe in \cite{5} determined optimal numbers of neurons in an MLPN for approximating the inverse kinematic function. To deal with complex manipulator structures some particular neural network architectures were presented, for example, a combination between the MLPN and the look-up table \cite{6}. A modular neural network in which the modules were concatenated in a global scheme in order to perform the inverse kinematics in a sequential way was proposed by Pablo \cite{7}. Similarly, in \cite{8,9,10} the inverse kinematic approximation using an RBFN was presented to compare with the performance of the MLPN. In this paper, the inverse function of the forward kinematic transformation is used to build the mapping from world coordinate space to joint angle space. In this paper two training methods, ACO and LMS were applied to train the RBFN. Training the RBFN using ACO represents a different idea from other existing papers \cite{8,9,10}. It is possible to determine an appropriate approximation of the inverse kinematics function. However, this solution has the main difficulty in how to collect accurate training patterns whose inputs are selected at pre-defined positions in the workspace of a real robotic system. Additionally, using the LMS to update the linear weights can improve the RBFN performance through online training. Therefore, the combination of ACO and LMS methods produces the advantages of both training methods to deal with the difficulty in collecting training patterns in practical applications. This approach consists of two steps, firstly producing an inaccurate inverse kinematics approximation by $\mathrm{ACO}$ and then recorrecting it through online training by the LMS. Based on the importance of RBF Neural Network center choices, a Radial Basis Function neural network learning algorithm based on ACO is presented to select a suitable basis function center. This algorithm can greatly improve the precision of the network and the training speed. The practical experiment and results are described in the concluding section which verifies the proposed approach.

%In summery, contributions of this paper are as follows,
%\begin{description}
%  \item[$\bullet$] We performed empirical study to understand interactions between disease-associated proteins and their direct neighbors.  
%   \item[$\bullet$] We develop a metric to measure connectedness of disease-associated proteins in a disease-assoaciated protein    
%\end{description}

\section{INVERSE KINEMATICS FUNCTION APPROXIMATION USING RADIAL BASIS FUNCTION NETWORK}  

The basic architecture of an RBFN is the three-layer network consisting of the input layer, hidden layer, and linear output layer \cite{11}. In the inverse kinematics problem, the inputs and outputs of the RBFN are position (image coordinates) and joint angles of the manipulator respectively. The unique feature of the RBFN compared to the MLP and other networks is the process performed at the hidden layer. In this hidden layer, the radial basis function works as a local selector in which the corresponding output depends on the distance between its center and input. It can be presented as:

\begin{equation} 
\phi_{i}(x) = w_i \exp\left ( -\frac{\left \| x-\mu_i \right \|^2}{2\sigma_i^{2}} \right )
\end{equation}

Where $\phi_{i}(x)$ is the radial basis function, $\mu_{i}$ is the central position of neuron $i$ and $\sigma_{i}$ is the width of the Gaussian function. The weightings, $w_{i}$ between the hidden layer and the output layer are adjusted based on an adaptive rule. $x$ is the input to the RBFN.

The total output of the RBFN is:

\begin{equation}
y_{j}(x)=\sum_{i=1}^{M} w_{j i} \phi_{i}(x)
\end{equation}

Where $w_{j i}$ is the weight between the $i$-th hidden unit and the $j$-th output.

\subsection{LEARNING METHOD OF THE NETWORK}

\begin{figure}
\includegraphics[width=0.5\textwidth]{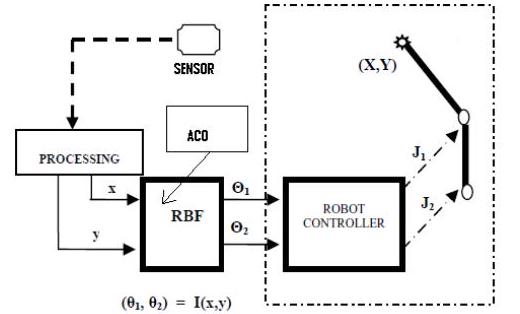}
\caption{ACO-RBFN control of robot manipulator} 
\end{figure}

Learning Algorithms of the proposed RBF network are composed of two parts.

\begin{itemize}
\item All samples of input used the ACO algorithm to identify the center and width of hidden layer neurons.
\item When the center is identified, we train the weight between the hidden layer and output layer. 
\end{itemize}

\subsection{DESIGN OF RBFN MODEL BASED ON ACO}

ACO was inspired by the nature of ant behavior. Ant individuals transmitted information through the volatile chemical substances which ants left in their passing path and also known as the "Pheromone" and then reached the purpose of finding the best way to search for food sources. When the later ants encountered pheromone, they not only can detect the presence and the number of the substance but also can guide their own choice of direction according to Pheromone concentration. Meanwhile, the material will be gradually volatile with the passage of time, so the length of the path and the number of ants that passed this path had an impact on the concentration of residual pheromone. For the same reason, the concentration size of residual pheromone guided the course of later ant action. So the more ants passed the path, the probability of the later ants choosing the path is greater. This is a message feedback phenomenon that is shown through the ant group behavior. ACO algorithm can achieve Intelligent Search, and global optimization, and has features of robustness, positive feedback, distributed computing, and being easily combined with other algorithms. The basic steps of the ACO algorithm are as follows: 
\\
(1) Set the number of iterations $N C=0$, set up the largest number of iterations $N C_{\max }$, initialize the pheromone $\tau_{i j}, \Delta \tau_{i j}^{k}(t)=0$, place $\mathrm{M}$ ants on $\mathrm{N}$ vertices.
\\
(2) Place the initial starting point of ants on current solution set. Every ant $\mathrm{k}$ according to the probability $p_{i j}^{k}(t)$ chose next vertex $\mathrm{j}$; place vertex $\mathrm{j}$ on current solution set. Path selection rules are defined as:

$$
p_{i j}^{k}(t)=\frac{\tau_{i j}(t)}{\sum_{j=1}^{N} \tau_{i j}(t)}
$$
\\
(3) Calculate the objective function value $z_{k}$ of each ant that records the best solution.
\\
(4) Set $t \leftarrow t+n ; N C \leftarrow N C+1$, update the pheromone of the path according to the updated rules of pheromone.

The update rule for global pheromone level is:
$$
\tau_{i j}(t+n)=(1-\rho) \tau_{i j}(t)+\rho \Delta \tau_{i j}(t)
$$
where $\rho \in(0,1)$ is a volatile factor. Pheromone increment $\Delta \tau_{i j}(t)$ can be expressed as:

$\Delta \tau_{i j}(t)=\sum_{k=1}^{M} Q / L_{k}$

Where $\mathrm{Q}$ is constant, $L_{k}$ is the path length that ant $\mathrm{k}$ in this cycle have walked.
\\
(5) If the ant group all converges to a path or cyclic times $N C \geq N C_{\max }$, then end the cycle and output the best path, otherwise repeat. The ACO based RBFN algorithm is described as follows:

(a) Use $\mathrm{ACO}$ algorithm to get the center $\mu_{i}$ $(i=1,2, \ldots, M)$ of RBF Neural Network ;

(b) Calculate the width of RBF according to the following formula:
$$
\sigma_{i}=d_{m} / \sqrt{2 M} \quad, \quad d_{m} \text { is the greatest distance }
$$
where $M$ is the number of the hidden nodes; 
\\
(c) Use the Least Mean Square algorithm to get the weight value $w_{j i}$ of the output layer. The error function of the network is
$$
E=\frac{1}{2} \sum_{k=1}^{j}\left(y_{k}-d_{k}\right)^{2}
$$
\\
(6) If $E>e$ ( $e$ is the precision that the network requires to achieve), then the number of hidden node increase by one, where the rule of weight adjustment is as follows:
$$
w_{j i}(t+1)=w_{j i}(t)+\alpha \frac{\partial E_{i}}{\partial w_{j i}}
$$
(7) Where $\alpha$ is a constant $0<\alpha<1$, repeating these steps until $E \leq e$. At this stage, the connection weights are to satisfy conditions.

\section{RESULTS}

The simulation result shows that the proposed method works considerably well in the presence of friction and external disturbance.
In this case, the proposed adaptive SMC is used on a three-link scara robot, with parameter matrices given by:

$$
\begin{aligned}
&M(q)=\left(\begin{array}{ccc}
M_{11} & M_{12} & 0 \\
M_{21} & M_{22} & 0 \\
0 & 0 & M_{33}
\end{array}\right), \\
&C(q, \dot{q})=\left(\begin{array}{ccc}
C_{11} & C_{12} & 0 \\
C_{21} & C_{22} & 0 \\
0 & 0 & 0
\end{array}\right) \quad G(q)=\left[\begin{array}{l}
0 \\
0 \\
G_{3}
\end{array}\right]
\end{aligned}
$$
Where

$M_{11}=l_{1}^{2}\left(\frac{m_{1}}{3}+m_{2}+m_{3}\right)+l_{1} l_{2}\left(m_{2}+2 m_{3}\right) \cos \left(q_{2}\right)+$

$l_{2}^{2}\left(\frac{m_{2}}{3}+m_{3}\right)$

$M_{13}=M_{23}=M_{31}=M_{32}=0$

$M_{12}=l_{1} l_{2}\left(\frac{m_{2}}{2}+m_{3}\right) \cos \left(q_{2}\right)-l_{2}^{2}\left(\frac{m_{2}}{3}+m_{3}\right)=M_{21}$ $M_{22}=l_{2}^{2}\left(\frac{m_{2}}{3}+m_{3}\right)$

$M_{33}=m_{3}$

$C_{1}=l_{1} l_{2} \sin \left(q_{2}\right)$

$C_{11}=-q_{2} C_{1}\left(m_{2}+2 m_{3}\right)$

$C_{12}=-q_{2} C_{1}\left(\frac{m_{2}}{2}+m_{3}\right)=C_{21}$

$C_{13}=C_{22}=C_{23}=C_{31}=C_{32}=C_{33}=0$

$G_{3}=-m_{3} g$

In which $q_{1}, q_{2}, q_{3}$ are the angle of joints 1,2 and 3 ; $m_{1}, m_{2}, m_{3}$ are the mass of the links 1,2 and 3 ; $l_{1}, l_{2}, l_{3}$ are the length of links 1,2 and 3 ; $g$ is the gravity acceleration. 

The system parameters of the scara robot are selected as the following:

$l_{1}=1.0 m ; l_{2}=0.8 m ; l_{3}=0.6 m$

$m_{1}=1.0 \mathrm{~kg} ; m_{2}=0.8 \mathrm{~kg} ; m_{3}=0.5 \mathrm{~kg} ;$

$g=9.8$ 

Important parameters that effect the control performance of the robotic system are the external disturbance $t_{1}(t)$, and friction term $f(q)$.

External disturbances are selected as:

$t_{1}(t)=\left[\begin{array}{l}5 \sin (2 t) \\ 5 \sin (2 t) \\ 5 \sin (2 t)\end{array}\right]$

Friction forces considered in these simulations as the following:

$f(q)=\left[\begin{array}{c}12 q_{1}+0.2 \operatorname{sign}\left(q_{1}\right) \\ 12 q_{2}+0.2 \operatorname{sign}\left(q_{2}\right) \\ 12 q_{3}+0.2 \operatorname{sign}\left(q_{3}\right)\end{array}\right]$

In this simulation the robot manipulator considered to carry a load of $10 \mathrm{~kg}$ to $20 \mathrm{~kg}$ with no prior knowledge of the weight; using 1 second to 4 seconds of the total simulation time.

The desired trajectories for the three joints to be tracked are given as follows:
$$
\begin{aligned}
&q_{d 1}(t)=1+0.1(\sin (t)+\sin (2 t)) \\
&q_{d 2}(t)=1+0.1(\cos (2 t)+\cos (3 t)) \\
&q_{d 3}(t)=1+0.1(\sin (3 t)+\sin (4 t))
\end{aligned}
$$
In this simulation, the model is estimated by applying a factor to the corresponding parameter matrices of the original system in each environment to count uncertainties, i.e.

$\hat{M}=0.9 M, \hat{C}=0.8 C, \hat{G}=0.85 G$

RBF parameters are $\varepsilon=0.05, k=15, \alpha(0)=0.5$. ACO parameters $\quad \rho=0.9, \varepsilon=1.0, Q=100$, $N C_{\max }=100$.

\begin{figure*}[hpbt]

     %\centering
     \subfloat[]{\includegraphics[width=0.35\textwidth]{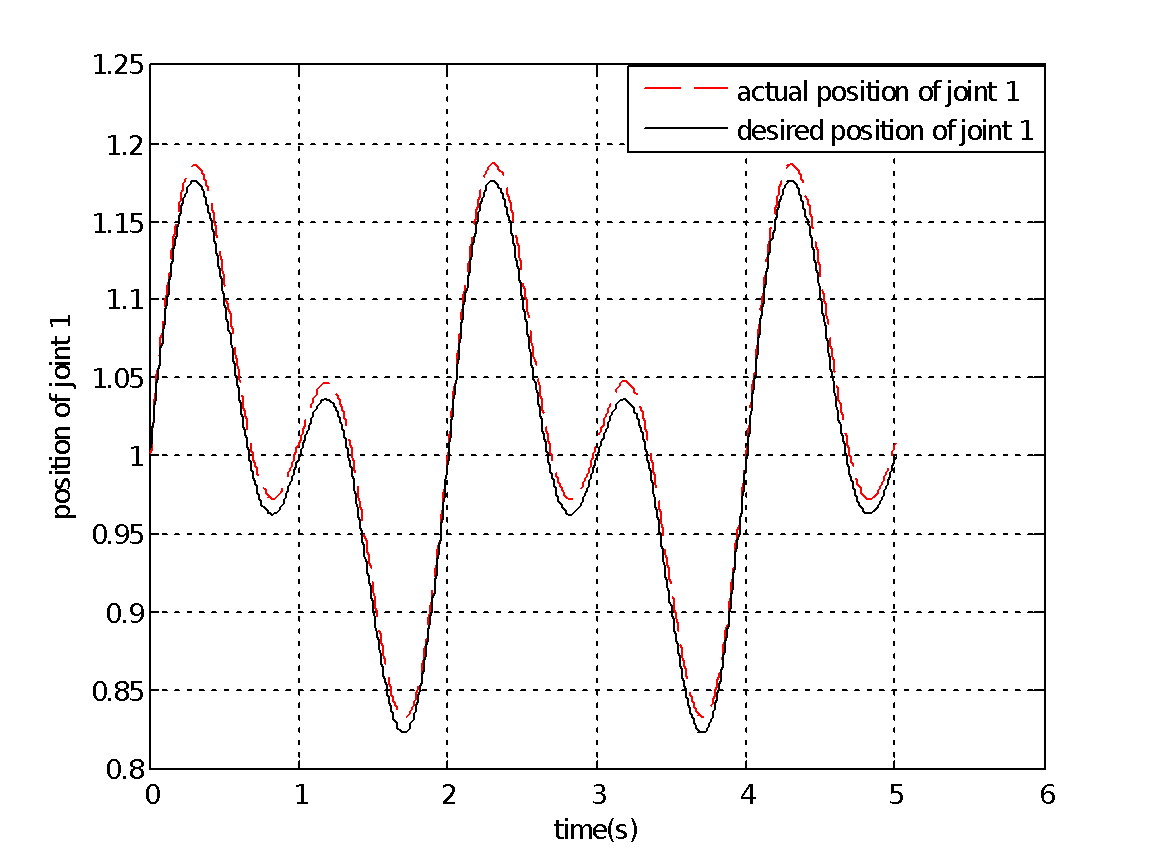}\label{figure1}}
     \subfloat[]{\includegraphics[width=0.35\textwidth]{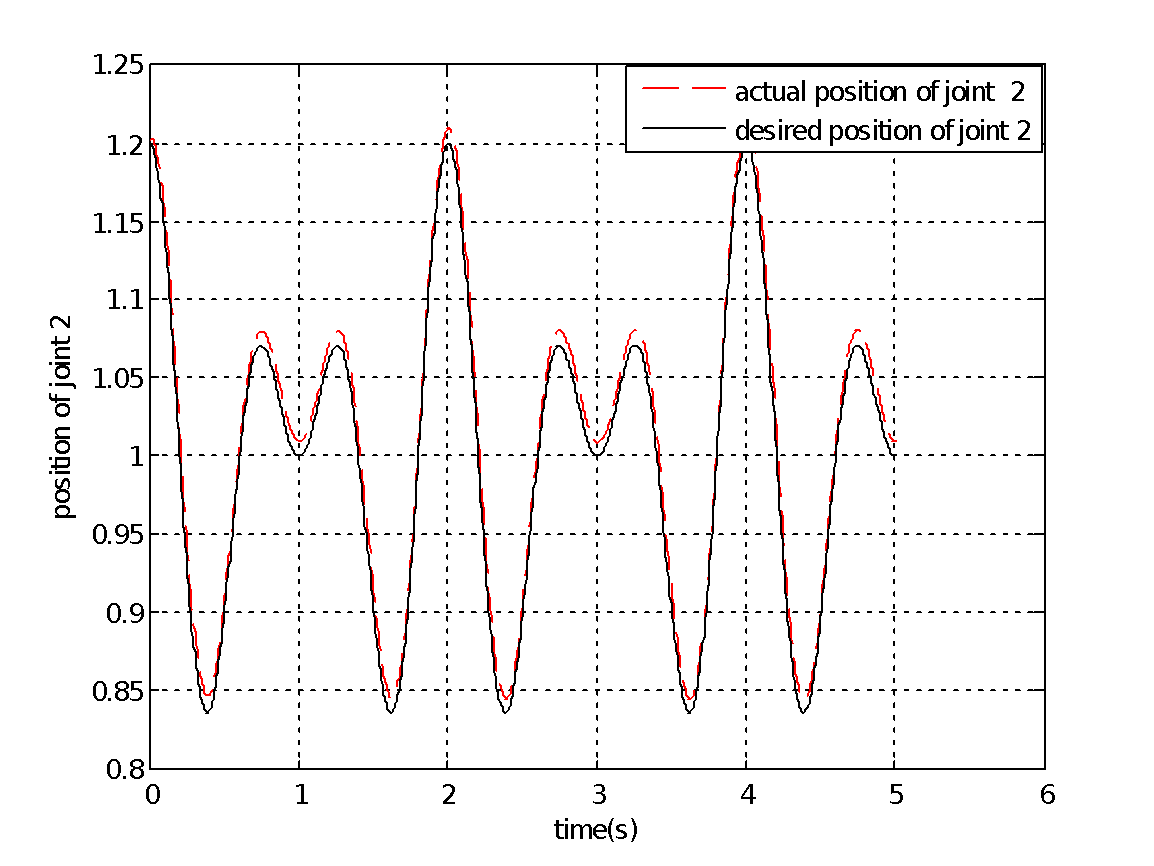}\label{figure2}}
     \subfloat[]{\includegraphics[width=0.35\textwidth]{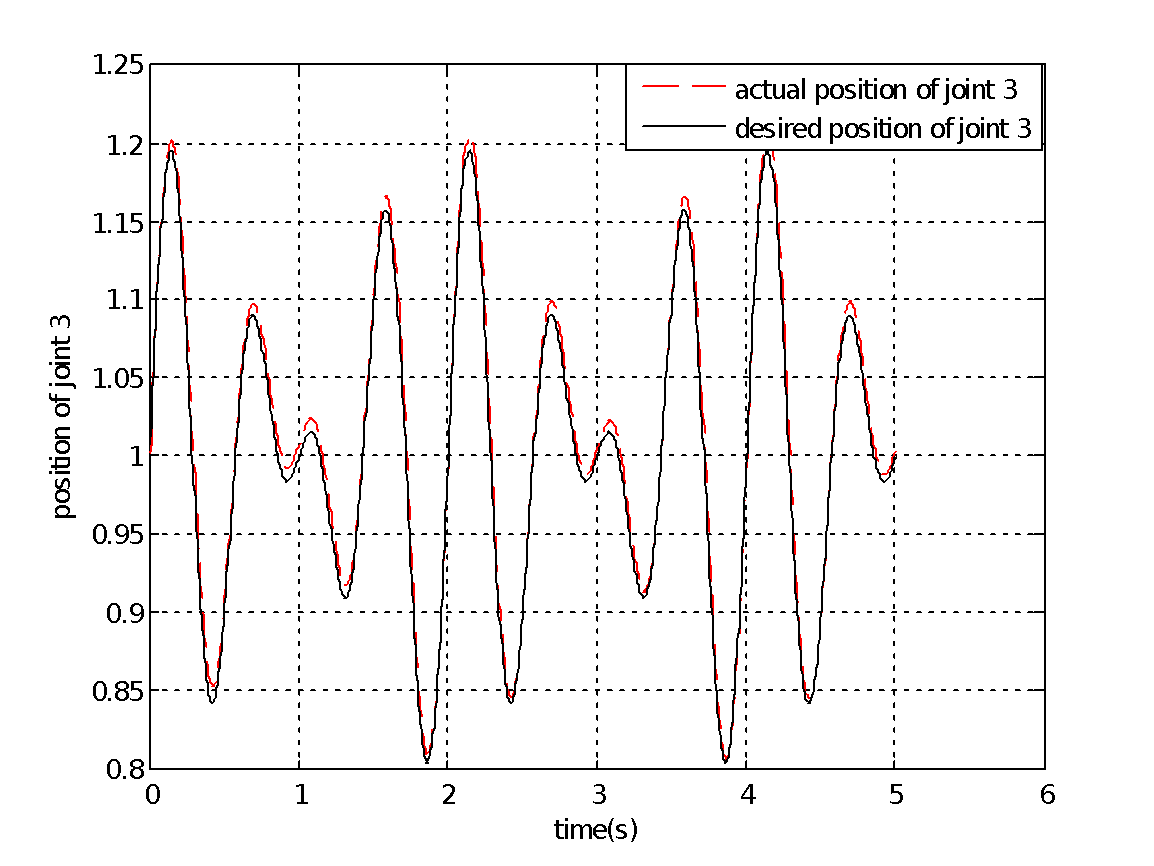}\label{figure3}}

     \subfloat[]{\includegraphics[width=0.35\textwidth]{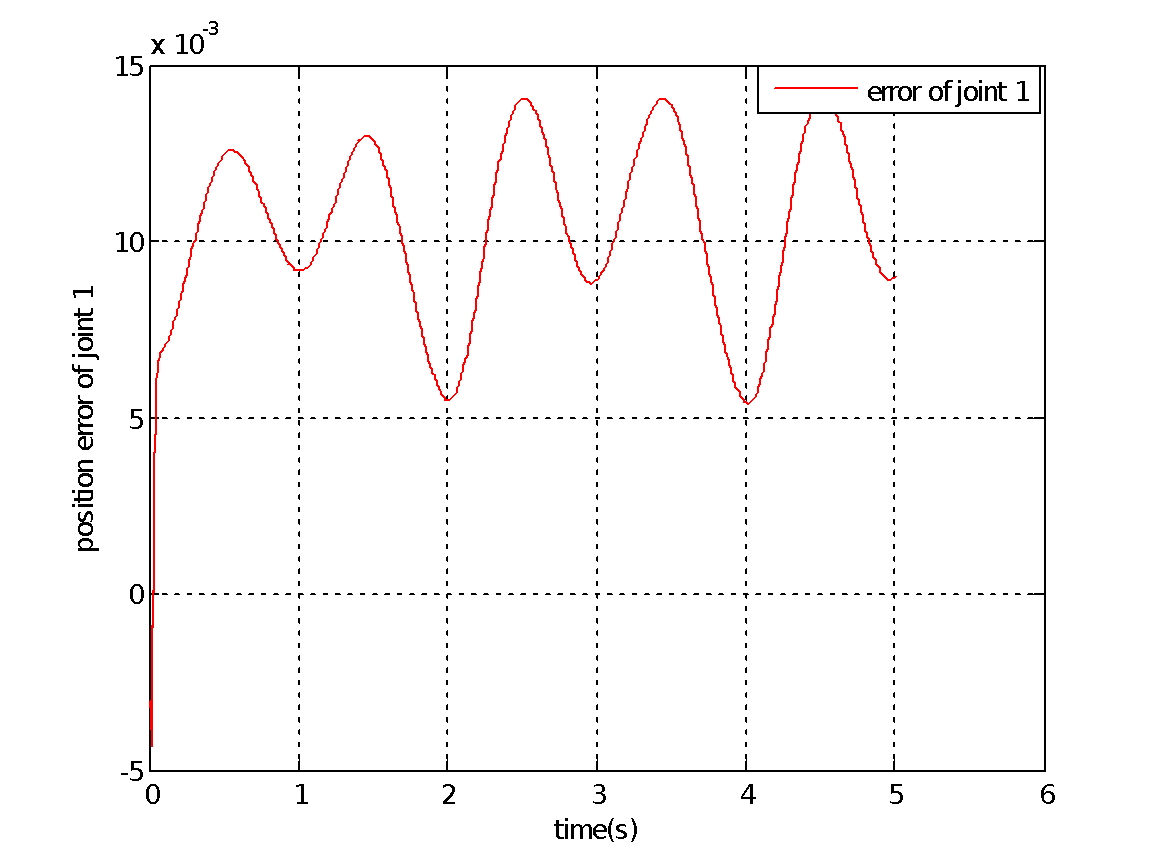}\label{figure4}}
     \subfloat[]{\includegraphics[width=0.35\textwidth]{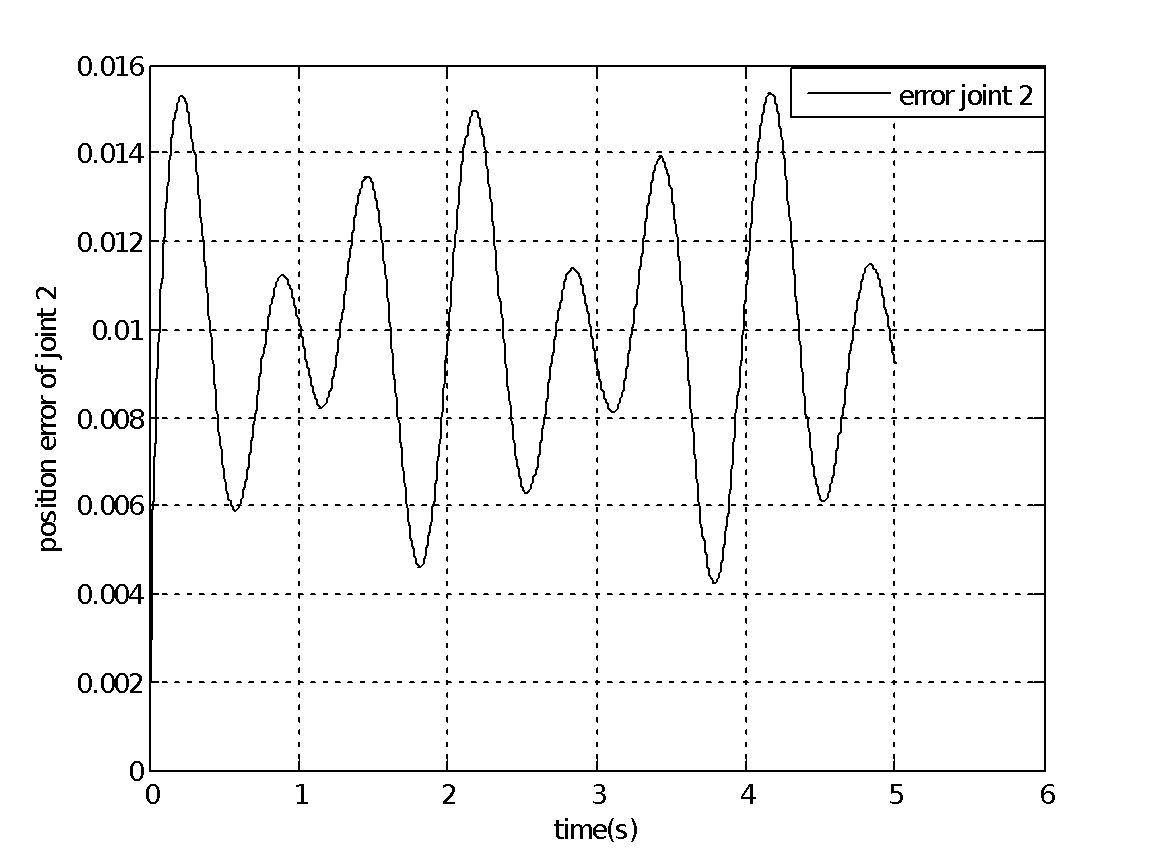}\label{figure5}}
     \subfloat[]{\includegraphics[width=0.35\textwidth]{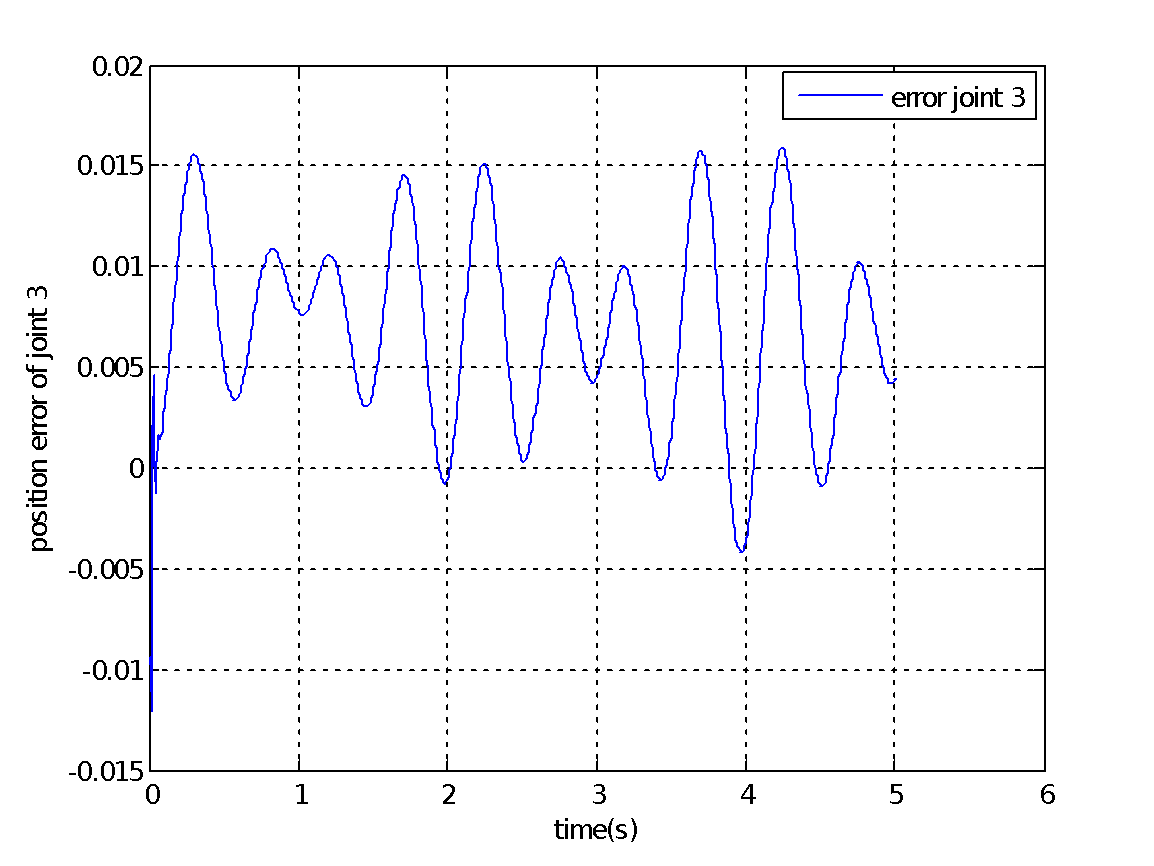}\label{figure6}}

     \subfloat[]{\includegraphics[width=0.35\textwidth]{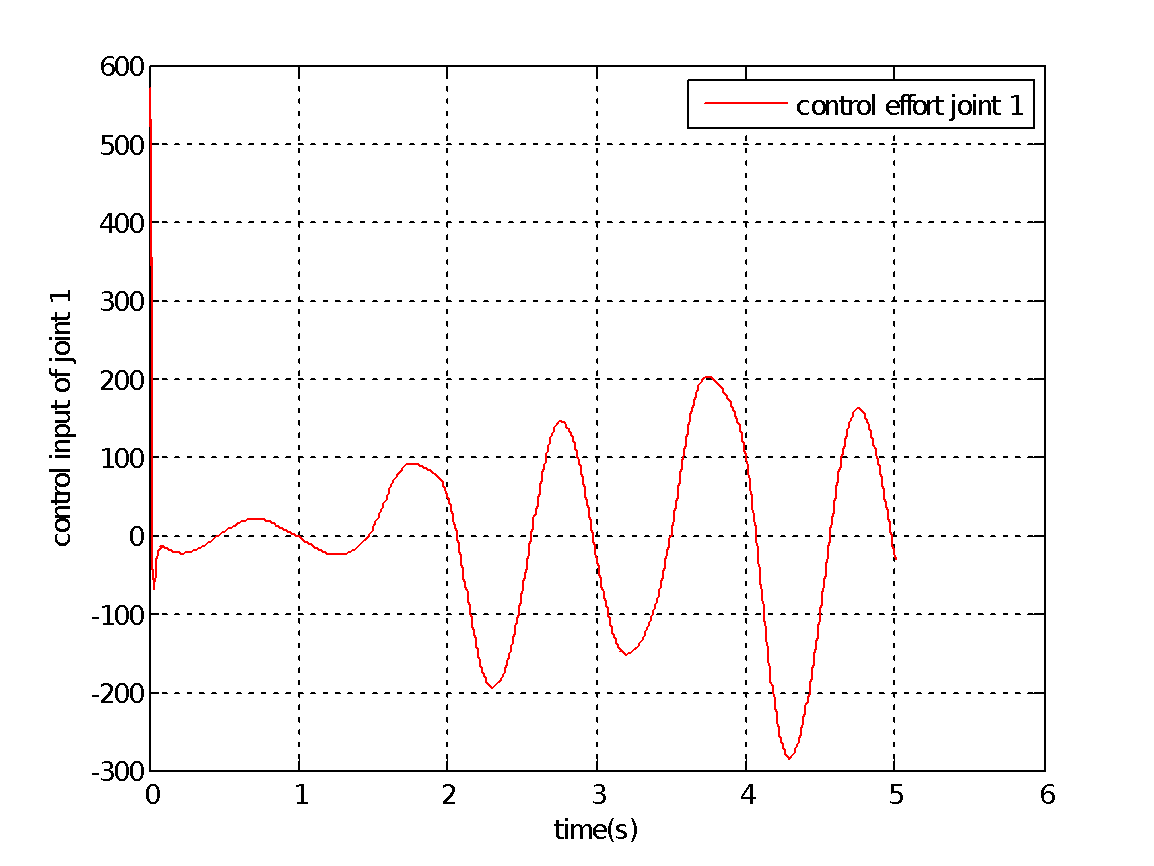}\label{figure7}}
     \subfloat[]{\includegraphics[width=0.35\textwidth]{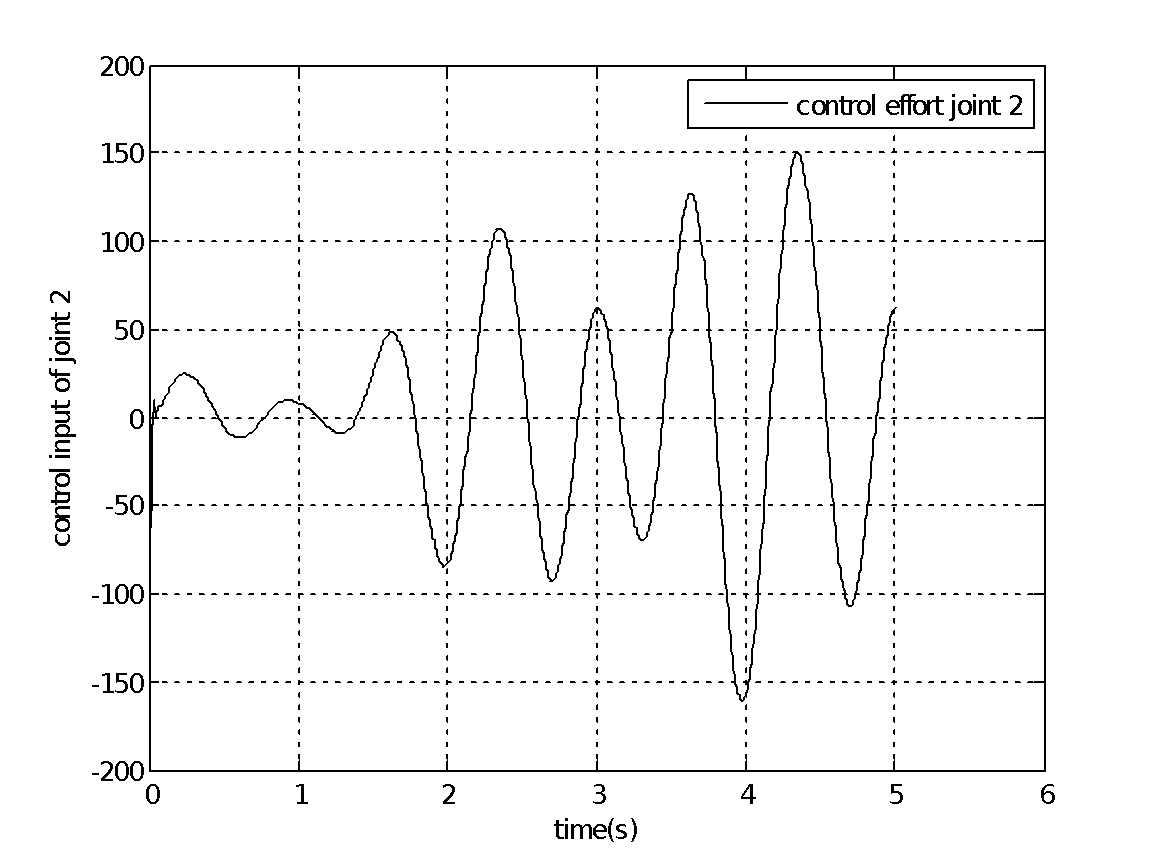}\label{figure8}}
     \subfloat[]{\includegraphics[width=0.35\textwidth]{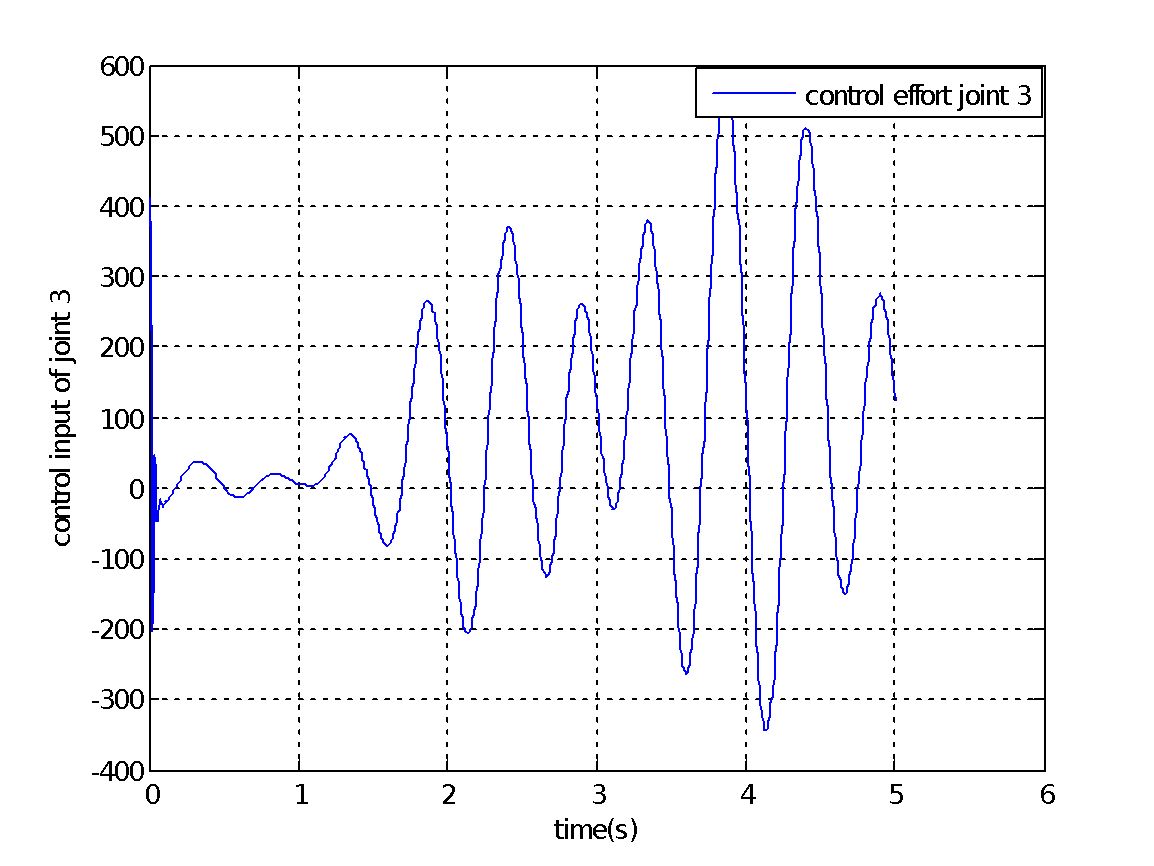}\label{figure9}}
	
     \caption{ Simulation results (a) position of joint 1 (b) position of joint 2 (c) position of joint 3 (d) position error of joint 1 (e) position error of joint 2 (f) position error of joint 3 (g) control input of joint 1 (h) control input of joint 2 (i) control input of joint 3.}
     \label{steady_state}
\end{figure*}

\section{CONCLUSION}
In this paper, an ACO based adaptive controller using RBF neural network is proposed for robotic manipulators. We used ACO to identify the center and width of hidden layer neurons. As shown in the experimental results, the proposed controller works considerably well in the presence of friction and external disturbance.

\end{document}